\pdfoutput=1
\documentclass{article}

\usepackage{microtype}
\usepackage{graphicx}
\usepackage{subcaption}
\usepackage{booktabs} 
\usepackage[url]{}
\usepackage{amsmath}
\usepackage{amssymb}
\usepackage{bbm}
\usepackage{gensymb}
\usepackage{amsmath,amssymb}

\usepackage[font=small,skip=0pt]{caption}
\usepackage{color, colortbl}
\definecolor{LightCyan}{rgb}{0.88,1,1}
\definecolor{aqua}{rgb}{0,1,1}

\usepackage{hyperref}

\usepackage[accepted]{icml2018}

\icmltitlerunning{Attend Before you Act: Leveraging human visual attention for continual learning}
\begin{document}

\twocolumn[
\icmltitle{Attend Before you Act: Leveraging human visual attention for continual learning}



\icmlsetsymbol{equal}{*}

\begin{icmlauthorlist}
\icmlauthor{Khimya Khetarpal}{to}
\icmlauthor{Doina Precup}{to,too}
\end{icmlauthorlist}

\icmlaffiliation{to}{Mila - McGill University, Montreal, Canada}
\icmlaffiliation{too}{Google Deepmind}

\icmlcorrespondingauthor{Khimya Khetarpal}{khimya.khetarpal@mail.mcgill.ca}
\icmlcorrespondingauthor{Doina Precup}{dprecup@cs.mcgill.ca}

\icmlkeywords{Machine Learning, ICML}

\vskip 0.3in
]



\printAffiliationsAndNotice{}  

\begin{abstract}
When humans perform a task, such as playing a game, they selectively pay attention to certain parts of the visual input, gathering relevant information and sequentially combining it to build a representation from the sensory data. In this work, we explore leveraging where humans look in an image as an implicit indication of what is salient for decision making. We build on top of the UNREAL architecture~\cite{jaderberg2016reinforcement} in DeepMind Lab's $3D$ navigation maze environment. We train the agent both with original images and foveated images, which were generated by overlaying the original images with saliency maps generated using a real-time spectral residual technique. We investigate the effectiveness of this approach in transfer learning by measuring performance in the context of noise in the environment.
\end{abstract}

\section{Introduction}
\label{submission}
Knowing where to look plays an important role in people's ability to learn and solve new tasks quickly. While some cues in an image are naturally attractive and lead to bottom-up saliency [\cite{harel2007graph}, \cite{walther2006modeling}], others need voluntary effort and are more task-dependent, leading to top-down saliency [\cite{sprague2004eye} \cite{borji2011cost}]. When humans perform a specific task, a combined model of attentional selection and object recognition is usually at work. Bottom-up feature extraction coupled with a hierarchical representation of object classes and motor commands governs subsequent eye movements in order to maximize  information gain \cite{itti2001computational}. 

The human attention mechanism is very complicated and depends on various factors ranging from task complexity, the nature of the task, external factors such as rewards or distractors, and internal factors such as curiosity. The work of \cite{triesch2003you} concluded that what we see is highly dependent on what we need. 
Human visual attention can also be seen as relying on a hierarchical approach \cite{baylis1993visual}. In particular, when performing a complex task which involves various subgoals, humans use selective attention to parts of the visual scene, sequential deployment of gaze in a temporal sequence of frames, before performing motor actions. More importantly, human attention reuses past understanding of concepts, relations, and world models. 

Intuitively, the fact that people focus only on specific parts of an image before acting should lead both to robustness in the presence of noise and deliberate distractors, as well as to the ability to generalize knowledge over different tasks. For example, we can navigate through any building regardless of the color of the walls or the interior decor. Hence, we would like to investigate if using this mechanism also provides robustness and ability to transfer knowledge for reinforcement learning (RL) agents as well.


Our goal is to explore how foveating around the regions where humans look impacts the reinforcement learning process, especially focusing on robustness and continual learning.  Because of this goal, we build on top of the UNREAL agent \cite{jaderberg2016reinforcement}, which aims to construct a better representation for continual learning, by focusing not only on learning the optimal value function for the given task, but also on optimizing several pseudo-rewards or auxiliary tasks.  We investigate the impact of overlaying the real image with a mask that is determined by a model of human attention.
We use the spectral residual saliency method \cite{hou2007saliency} to foveate around salient regions and train the UNREAL agents on a $3D$ maze navigation task from DeepMind Lab \cite{beattie2016deepmind}. We use varying degrees of foveation, in order to evaluate the impact on the learning process.  Our hypothesis was that more foveation should lead to more robustness to distractors and noise, but also to worse final task performance. We also empirically explore if knowing where to look facilitates continual learning and leads learnt policies to be robust to variations in the data distribution.

\begin{figure}[h]
   \vskip 0.2in
  \begin{center}
  \centerline{\includegraphics[width=\linewidth]{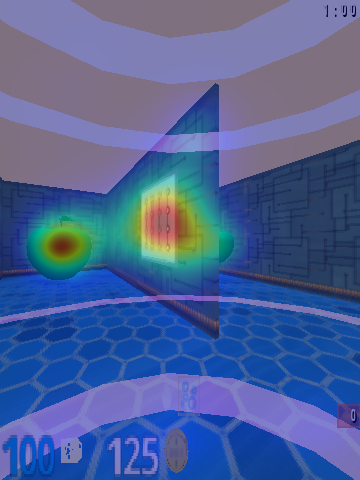}}
   \caption{\textbf{Sample image} from $3D$ navigation maze environment in DeepMind Lab overlaid with a saliency heat map generated from the pre-trained model of \citeauthor{cornia2016predicting}.}
   \label{fig:SaliencyMap}
   \end{center}
  \vskip -0.2in
\end{figure}

\section{Algorithmic approach}
We started our approach by investigating saliency maps generated from the state-of-the-art Saliency Attentive Model (SAM) \cite{cornia2016predicting} according to the MIT Saliency Benchmark \cite{bylinskii2015saliency}. Figure \ref{fig:SaliencyMap} shows a sample input image from a static maze navigation task overlaid with a heat map generated from SAM. SAM uses a Convolutional LSTM to focus on specific parts of the image and iteratively refines the visual attention. Once a gray scale saliency map is generated from SAM, we overlay it on the original image using \textit{jet} color map. More salient regions in the image are indicated by the hotness of the map i.e. the red color, whereas relatively insignificant regions are indicated by coolness of the map i.e. the blue color. While the saliency maps generated by SAM look very intuitive, using a SAM model pre-trained on the VGG dataset is computationally very expensive in terms of speed of training. For faster training, instead of SAM, we decided to use a real-time saliency computation technique called the Spectral Residual method \cite{hou2007saliency}. The key idea of this method is to compute the average frequency domain and subtract it from a specific image domain to obtain the spectral residual. The log spectrum of each image is analyzed to obtain the spectral residual, then it is transformed to a spatial domain with the location of the proto-objects. Proto-objects are pre-attentive structures with limited spatial and temporal coherence within a visual stimuli, which generate the perception of an object when attended to.
\begin{figure}[h]
	\begin{center}
		\begin{subfigure}[b]{0.2\textwidth}
			\centering
			\captionsetup{justification=centering}
			\includegraphics[width=\textwidth]{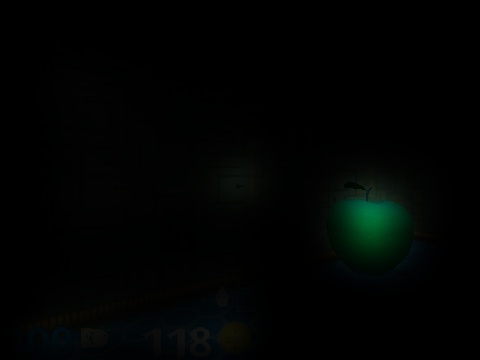}
			\caption[]%
			{{\small $\alpha=0$ }}    
			\label{fig:salientpartsonly}
		\end{subfigure}
		\quad
		\begin{subfigure}[b]{0.2\textwidth}  
			\centering
			\captionsetup{justification=centering}
			\includegraphics[width=\textwidth]{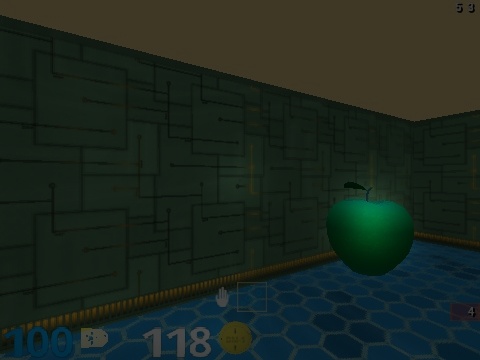}
			\caption[]%
			{{\small $\alpha=0.25$ }}    
			\label{fig:0.25alpha}
		\end{subfigure}
		\vskip\baselineskip
		\begin{subfigure}[b]{0.2\textwidth}   
			\centering
			\captionsetup{justification=centering}
			\includegraphics[width=\textwidth]{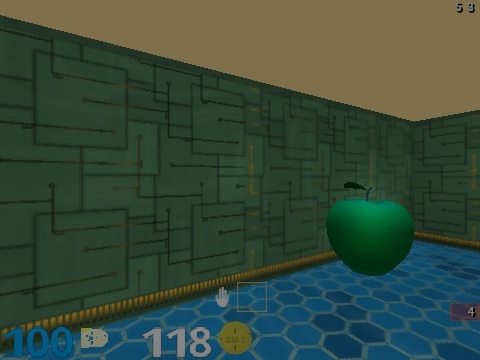}
			\caption[]%
			{{\small $\alpha=0.50$ }}    
			\label{fig:0.50alpha}
		\end{subfigure}
		\quad
		\begin{subfigure}[b]{0.2\textwidth}   
			\centering 
			\captionsetup{justification=centering}
			\includegraphics[width=\textwidth]{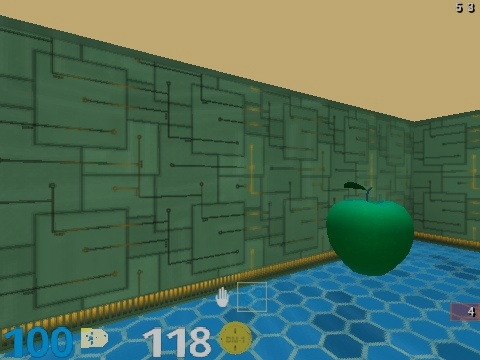}
			\caption[]%
			{{\small $\alpha=0.75$ }}
            \label{fig:0.75alpha}
		\end{subfigure}
		\caption[]
		{\small \textbf{Leveraging different degrees of foveation around where humans look in an image}: Different degrees of attention indicated by the parameter $\alpha$ signify the importance of salient regions with respect to the original image. We consider a full range from only looking at salient regions in Figure \ref{fig:salientpartsonly} to focus relatively more on the whole image in Figure \ref{fig:0.75alpha}. Saliency Attentive Model (SAM) is used to compute the saliency maps in this figure.} 
		\label{fig:varyingalpha}
	\end{center}
\end{figure}    

We first explore if foveating around the salient locations in the image helps the agent to learn faster. It is natural for humans to look at an entire visual scene, yet, automatically focus around salient regions while eliminating others which are not so important. With this intuition, instead of explicitly providing the attention map along with the original image, we blend the attention map with the original image, as follows:
\begin{equation}\label{blending}
I(x,y) =  I(x,y) +\Big( S(x,y) + \alpha (1 - S(x,y)) \Big),
\end{equation}
where $S(x,y)$ is the normalized saliency map for all pixels $(x,y)$, $I(x,y)$ denotes the original image, and $\alpha$ is the amount of foveation, and controls the amount of blending desired. This is also depicted in Figure \ref{fig:varyingalpha} for $\alpha$ ranging from $0$ to $1$. For instance, a value of $\alpha = 0$ indicates removing all distractors and focusing on salient regions alone (Figure \ref{fig:salientpartsonly}), whereas a value of $\alpha = 0.75$ implies looking largely at the original image.

We train the UNREAL agent on DeepMind Lab's \cite{beattie2016deepmind} static navigation maze task (nav maze static $01$) with all auxiliary tasks on as our baseline. We keep the network architecture consistent with the \citeauthor{jaderberg2016reinforcement}, with a CNN-LSTM base agent trained on-policy with A3C \cite{mnih2016asynchronous}. The input to the agent at each timestep was an $84$ * $84$ RGB image. The network consists of two convolutional layers with $16, 8*8$ and $32, 4*4$ filters respectively. This is followed by a fully connected layer with $256$ units. RELU activation function is used for all three layers. An LSTM is used with the inputs concatenated from the fully connected layer, previous action taken, and previous reward. Three auxiliary tasks include the pixel-control task, value-function replay and the reward prediction task as described in \citep{jaderberg2016reinforcement}. We use $20$ timestep rollouts for the base process. After every $20$ environment steps, the auxiliary tasks are performed corresponding to every update of the base A3C agent. We used the online open source-code of UNREAL\footnote{\url{https://github.com/miyosuda/unreal}} as our baseline.

Next, we introduce the Visually-Attentive UNREAL agent \footnote{The source code is available at \url{https://github.com/kkhetarpal/unrealwithattention}} by foveating around the salient regions in each image. This is done in the base process of online A3C , as shown in the pseudo code in Algorithm \ref{alg:visuallyattentiveUNREAL}. 

{
	\begin{algorithm}[h]
		\caption{Visually Attentive UNREAL Agent}
		\label{alg:visuallyattentiveUNREAL}
		\begin{algorithmic}
			\STATE   $\alpha$ is factor controlling the foveation
			\STATE $I \gets$ Obtain original Input Image of $360*480$ from the Lab environment
            \STATE $S \gets$ SpectralSaliencyMethod ($I$)
            \STATE $Foveated Image \gets$ SaliencyOverlay ($I$, $S$, $\alpha$)
            \STATE Process Base A3C CNN-LSTM (Foveated Image) 
            \STATE Process Auxiliary Tasks (Foveated Image)
		\end{algorithmic}
	\end{algorithm}
}

\begin{figure}[h]
   \vskip 0.2in
  \begin{center}
  \centerline{\includegraphics[width=1\linewidth]{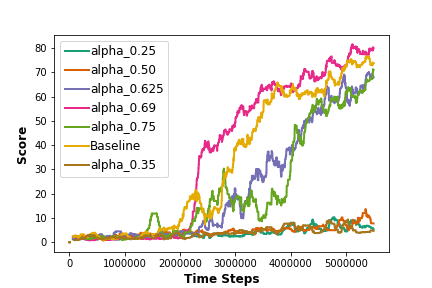}}
   \caption{\textbf{Learning with varying degrees of visual attention} to navigate the $3D$ maze environment. Specific degrees of visual attention helps in learning better than baseline UNREAL agent. Here $\alpha = 0.69$ speeds up the learning as compared to other settings for this instance of runs.}
   \label{AlphaLearningCurves}
   \end{center}
  \vskip -0.2in
\end{figure}

The training used $8$ parallel threads for all our experiments. For our preliminary experiments, we explored different values of $\alpha$. We can observe that on one hand, foveating on the salient regions alone removes a lot of context from the important aspects of an image and results in little to no learning, as seen in the Figure \ref{AlphaLearningCurves}. This is also intuitive from the visualization in the Figure \ref{fig:salientpartsonly}. On the other hand, values of $\alpha$ in the range of $(0.62, 0.70)$ show a boost in performance in the preliminary learning curves, as shown in Figure \ref{AlphaLearningCurves}.

\section{Experiments}

\begin{table*}[t]
\centering
\caption{\textbf{Transfer Learning :} Average Performance over $k=25$ games once training is completed. UNREAL agent and Visually-Attentive UNREAL agent are evaluated once training is stopped and also for transfer learning. Transfer is evaluated on three variations of training categorized as - \textbf{Easy:} Simple Gaussian noise is added in the original frames, \textbf{Moderate:} Tinting of frames is done by randomly flipping a coin with the same hue of 0.25, and \textbf{Difficult:} At random, some frames are tinted with different amounts of hue ranging from $0$ to $1$. Scores here are averaged for 25 games with standard deviation across these games in the brackets.
\label{finalscores}} 
\vskip 0.15in
\begin{tabular}{||c || c|| c|| c|| c||} 
 \hline
\textbf{Agent} &  \textbf{Testing}  &  \multicolumn{3}{|c|}{\textbf{Continual Learning}} \\ 
 \hline
  &  & Easy & Moderate & Difficult \\ 
 \hline
UNREAL & $96.92 (8.08)$ & $101.96  (9.656) $ & $92.64 (12.35)$
 & $39.16 (11.44)$ \\ 
 \hline
Visually-Attentive UNREAL & $95.92 (10.88)$ & $96.96 (9.39)
$ & $83.52 (10.09)$ & $40.52 ( 14.67)$ \\
 \hline\hline
 \end{tabular}
 \vskip -0.1in
\end{table*}

\begin{figure}[h]
   \vskip 0.2in
  \begin{center}
  \centerline{\includegraphics[width=1\linewidth]{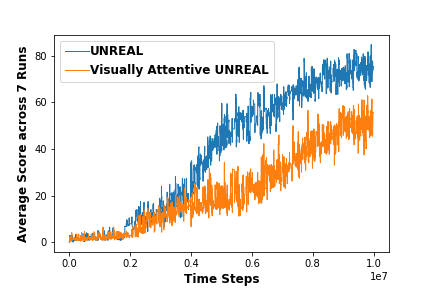}}
   \caption{\textbf{Learning curves} in the $3D$ Navigation Maze Static. On an average, UNREAL agent learns better than Visually-Attentive UNREAL agent during the training phase. However, both agents suffer from a lot of variance during training phase.}
   \label{fig:LearningCurves}
   \end{center}
  \vskip -0.2in
\end{figure}

Based on our preliminary results, we further trained the Visually-attentive UNREAL agents only using the value of $\alpha=0.69$ ,  which showed a boost in performance in Figure \ref{AlphaLearningCurves}. We ran multiple runs for both the  baseline and visually-attentive agent. Figure~\ref{fig:LearningCurves} shows the learning curves for $10M$ time steps averaged across $7$ runs. The Visually-Attentive UNREAL agent learns marginally slower than the baseline on an average. Moreover, the amount of foveation determines the impact on the learning. However, the learning curve only suggests how these agents perform in the same environment over time. Next we explore, how visually-attentive agents compare to the baseline in transfer of learning. In other words, \textit{does visual attention facilitate continual learning?}.

To evaluate the trained models for continual learning, we introduce three types of perturbations in the input frames and the average performance over \textit{k=25} games is recorded. Table~\ref{finalscores} depicts the performance averaged over $25$ games for both these agents. These variations include addition of Gaussian noise, tinting of images at random with the same hue, and tinting of images at random with different hues, categorized as three levels of difficulty namely easy, medium and hard. To tint the frames, we generate a flickering effect in the sequence of frames by scaling RGB values and by adjusting colors in the HSV color-space. From the mean scores in Table \ref{finalscores} one can note that both baseline and the visually attentive UNREAL agent remain unaffected in performance by relatively small amounts of Gaussian noise. Upon encountering flickering in frames at random, the visually-attentive UNREAL agent is still able to perform as well as the baseline and is relatively more robust to distractors in both easy and moderate categories of evaluation. However, both agents struggle to perform transfer learning when the amount of distraction is larger than what they have seen during training. For a qualitative analysis, we present the visualization of both these agents in all three test-scenarios as additional results in the supplementary material\footnote{\url{https://sites.google.com/view/attendbeforeyouact}}. 


\section{Discussion and Future Work}
\label{conclusionandfuturework}
We present an exploratory study to understand the role of visual attention in learning to perform a task and evaluating its effect in continual learning. Our key hypothesis is that knowing where to look in an image helps in learning a task, because this knowledge could be transferred to new tasks. We train the visually-attentive UNREAL agent which foveates around regions of an image salient to the human eye. The performance evaluation on perturbations in the train setting demonstrate promising results for further analysis of continual learning with visual attention.

\begin{figure}[h]
	\begin{center}
		\begin{subfigure}[b]{0.2\textwidth}
			\centering
			\captionsetup{justification=centering}
			\includegraphics[width=\textwidth]{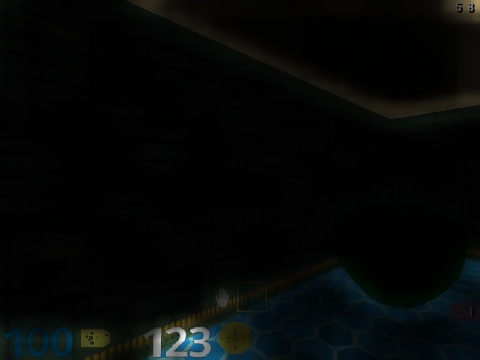}
			\caption[]%
			{{\small $\alpha=0$ }}    
			\label{fig:spectralsalientpartsonly}
		\end{subfigure}
		\quad
		\begin{subfigure}[b]{0.2\textwidth}  
			\centering
			\captionsetup{justification=centering}
			\includegraphics[width=\textwidth]{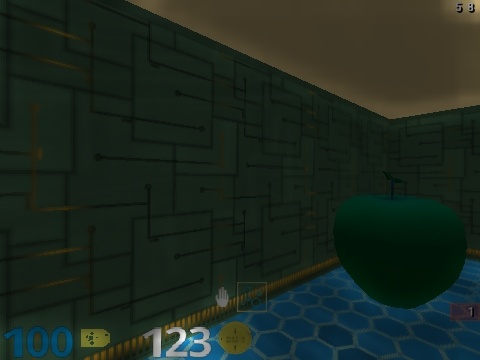}
			\caption[]%
			{{\small $\alpha=0.25$ }}    
			\label{fig:spectral0.25alpha}
		\end{subfigure}
		\vskip\baselineskip
		\begin{subfigure}[b]{0.2\textwidth}   
			\centering
			\captionsetup{justification=centering}
			\includegraphics[width=\textwidth]{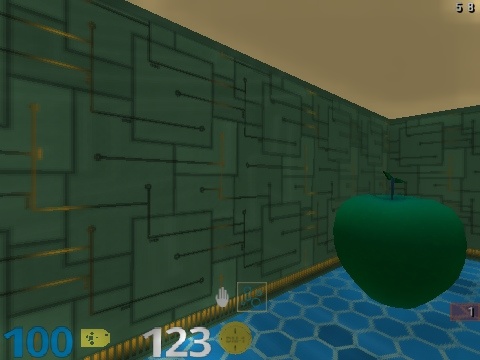}
			\caption[]%
			{{\small $\alpha=0.50$ }}    
			\label{fig:spectral0.50alpha}
		\end{subfigure}
		\quad
		\begin{subfigure}[b]{0.2\textwidth}   
			\centering 
			\captionsetup{justification=centering}
			\includegraphics[width=\textwidth]{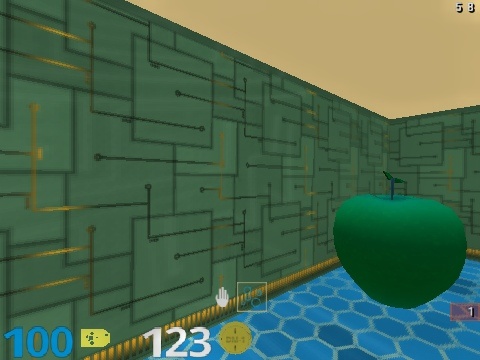}
			\caption[]%
			{{\small $\alpha=0.75$ }}
            \label{fig:spectral0.75alpha}
		\end{subfigure}
		\caption[]
		{\small \textbf{Qualitative analysis of Spectral Residual Method:}  Attention maps computed from the spectral residual method are used to generate different degrees of foveation indicated by the parameter $\alpha$. This model attends a lot more to the score than the other objects in the image.} 
		\label{fig:spectralvaryingalpha}
	\end{center}
 	\vskip -0.2in
\end{figure}

In this work, we employed a fundamental spectral residual saliency method which is based on the log spectra representation of images. However, this technique does not take into account the motion features which could be a limiting factor in terms of performance of the visually-attentive agent. This was further confirmed by qualitative analysis of the attention maps generated by the spectral residual saliency method as shown in the Figure \ref{fig:spectralvaryingalpha}. It is interesting to note that this model focuses a lot more on the score region of the frame than the objects in the maze. One of the potential reasons for limited performance is that computed attention maps focus on one most important object in the frame as opposed to all salient regions. We note that our approach can be used as a wrapper around {\em any} saliency model, so it would be easy to try better approaches.

A possible future direction in understanding the role of attention could involve training saliency models explicitly for images encountered in game playing. Even using pre-trained SAM model in an optimized fashion would potentially impact the performance. One could employ a better saliency model to help the agent foveate on regions which capture the dynamics of the rewards and the feature-representation. More importantly, it would be interesting to study a setting where the agent can actively learn to control where to attend to, rather than using a static attention model. \citeauthor{jayaraman2016look} 's work in learning object representation in a dynamic interactive setting relates to similar line of thought. Thus, an open question remains: how can we ensure that an agent controls the visit to the most visually attended states?

\bibliography{main}

\begin{thebibliography}{14}
\providecommand{\natexlab}[1]{#1}
\providecommand{\url}[1]{\texttt{#1}}
\expandafter\ifx\csname urlstyle\endcsname\relax
  \providecommand{\doi}[1]{doi: #1}\else
  \providecommand{\doi}{doi: \begingroup \urlstyle{rm}\Url}\fi

\bibitem[Baylis \& Driver(1993)Baylis and Driver]{baylis1993visual}
Baylis, Gordon~C and Driver, Jon.
\newblock Visual attention and objects: evidence for hierarchical coding of
  location.
\newblock \emph{Journal of Experimental Psychology: Human Perception and
  Performance}, 19\penalty0 (3):\penalty0 451, 1993.

\bibitem[Beattie et~al.(2016)Beattie, Leibo, Teplyashin, Ward, Wainwright,
  K{\"u}ttler, Lefrancq, Green, Vald{\'e}s, Sadik, et~al.]{beattie2016deepmind}
Beattie, Charles, Leibo, Joel~Z, Teplyashin, Denis, Ward, Tom, Wainwright,
  Marcus, K{\"u}ttler, Heinrich, Lefrancq, Andrew, Green, Simon, Vald{\'e}s,
  V{\'\i}ctor, Sadik, Amir, et~al.
\newblock Deepmind lab.
\newblock \emph{arXiv preprint arXiv:1612.03801}, 2016.

\bibitem[Borji et~al.(2011)Borji, Ahmadabadi, and Araabi]{borji2011cost}
Borji, Ali, Ahmadabadi, Majid~N, and Araabi, Babak~N.
\newblock Cost-sensitive learning of top-down modulation for attentional
  control.
\newblock \emph{Machine Vision and Applications}, 22\penalty0 (1):\penalty0
  61--76, 2011.

\bibitem[Bylinskii et~al.(2015)Bylinskii, Judd, Borji, Itti, Durand, Oliva, and
  Torralba]{bylinskii2015saliency}
Bylinskii, Zoya, Judd, Tilke, Borji, Ali, Itti, Laurent, Durand, Fr{\'e}do,
  Oliva, Aude, and Torralba, Antonio.
\newblock Mit saliency benchmark, 2015.

\bibitem[Cornia et~al.(2016)Cornia, Baraldi, Serra, and
  Cucchiara]{cornia2016predicting}
Cornia, Marcella, Baraldi, Lorenzo, Serra, Giuseppe, and Cucchiara, Rita.
\newblock Predicting human eye fixations via an lstm-based saliency attentive
  model.
\newblock \emph{arXiv preprint arXiv:1611.09571}, 2016.

\bibitem[Harel et~al.(2007)Harel, Koch, and Perona]{harel2007graph}
Harel, Jonathan, Koch, Christof, and Perona, Pietro.
\newblock Graph-based visual saliency.
\newblock In \emph{Advances in neural information processing systems}, pp.\
  545--552, 2007.

\bibitem[Hou \& Zhang(2007)Hou and Zhang]{hou2007saliency}
Hou, Xiaodi and Zhang, Liqing.
\newblock Saliency detection: A spectral residual approach.
\newblock In \emph{Computer Vision and Pattern Recognition, 2007. CVPR'07. IEEE
  Conference on}, pp.\  1--8. IEEE, 2007.

\bibitem[Itti \& Koch(2001)Itti and Koch]{itti2001computational}
Itti, Laurent and Koch, Christof.
\newblock Computational modelling of visual attention.
\newblock \emph{Nature reviews neuroscience}, 2\penalty0 (3):\penalty0 194,
  2001.

\bibitem[Jaderberg et~al.(2016)Jaderberg, Mnih, Czarnecki, Schaul, Leibo,
  Silver, and Kavukcuoglu]{jaderberg2016reinforcement}
Jaderberg, Max, Mnih, Volodymyr, Czarnecki, Wojciech~Marian, Schaul, Tom,
  Leibo, Joel~Z, Silver, David, and Kavukcuoglu, Koray.
\newblock Reinforcement learning with unsupervised auxiliary tasks.
\newblock \emph{arXiv preprint arXiv:1611.05397}, 2016.

\bibitem[Jayaraman \& Grauman(2016)Jayaraman and Grauman]{jayaraman2016look}
Jayaraman, Dinesh and Grauman, Kristen.
\newblock Look-ahead before you leap: end-to-end active recognition by
  forecasting the effect of motion.
\newblock In \emph{European Conference on Computer Vision}, pp.\  489--505.
  Springer, 2016.

\bibitem[Mnih et~al.(2016)Mnih, Badia, Mirza, Graves, Lillicrap, Harley,
  Silver, and Kavukcuoglu]{mnih2016asynchronous}
Mnih, Volodymyr, Badia, Adria~Puigdomenech, Mirza, Mehdi, Graves, Alex,
  Lillicrap, Timothy, Harley, Tim, Silver, David, and Kavukcuoglu, Koray.
\newblock Asynchronous methods for deep reinforcement learning.
\newblock In \emph{International Conference on Machine Learning}, pp.\
  1928--1937, 2016.

\bibitem[Sprague \& Ballard(2004)Sprague and Ballard]{sprague2004eye}
Sprague, Nathan and Ballard, Dana.
\newblock Eye movements for reward maximization.
\newblock In \emph{Advances in neural information processing systems}, pp.\
  1467--1474, 2004.

\bibitem[Triesch et~al.(2003)Triesch, Ballard, Hayhoe, and
  Sullivan]{triesch2003you}
Triesch, Jochen, Ballard, Dana~H, Hayhoe, Mary~M, and Sullivan, Brian~T.
\newblock What you see is what you need.
\newblock \emph{Journal of vision}, 3\penalty0 (1):\penalty0 9--9, 2003.

\bibitem[Walther \& Koch(2006)Walther and Koch]{walther2006modeling}
Walther, Dirk and Koch, Christof.
\newblock Modeling attention to salient proto-objects.
\newblock \emph{Neural networks}, 19\penalty0 (9):\penalty0 1395--1407, 2006.

\end{thebibliography}
\bibliographystyle{icml2018}
\end{document}